\documentclass[letterpaper, 10 pt, conference]{ieeeconf} 
\usepackage[utf8]{inputenc}

\IEEEoverridecommandlockouts
\usepackage{units}
\usepackage{algorithm, algorithmic}
\usepackage{amsmath}
\usepackage{amssymb}
\usepackage{bm}
\usepackage{tabularx}
\usepackage{booktabs}
\usepackage{cite} 
\usepackage{graphicx}
\usepackage{subcaption}
\usepackage{caption}
\usepackage{mwe}
\usepackage{hyperref}
\usepackage[svgnames]{xcolor}
\usepackage{multirow}
\usepackage{makecell}
\usepackage{xcolor,colortbl}
\usepackage{svg}
\hypersetup{
 colorlinks = true,
 linkcolor=black,
 urlcolor=Blue}

\usepackage{amsthm} 

\newif\ifcorrectingmode
\correctingmodefalse
\usepackage[normalem]{ulem}
\ifcorrectingmode
	
	\newcommand{\deleted}[1]{\textcolor{Black}{\ifmmode\text{\sout{\ensuremath{#1}}}\else\sout{#1}\fi}}
	\newcommand{\deletedequation}[2]{\textcolor{Black}{\centerline{Removed equation (#1)}}}
\else
	
	\newcommand{\deleted}[1]{}
	\newcommand{\deletedequation}[2]{}
\fi

\usepackage[printonlyused]{acronym}
\newacro{DOF}{degree of freedom}
\newacroplural{DOF}{degrees of freedom}
\acrodefplural{DOF}[DOFs]{degrees of freedom}
\newacro{iLQR}{Iterative Linear-Quadratic Regulator}
\newacro{CT}{Control Toolbox}
\newacro{EOM}{equations of motion}
\newacro{OC}{Optimal Control}
\newacro{LQR}{linear-quadratic regulator}
\newacro{PD}{proportional derivative}
\newacro{MPC}{Model Predictive Control}
\newacro{LQ}{linear quadratic}
\newacro{LQOC}{Linear-Quadratic Optimal Control}
\newacro{TO}{Trajectory Optimization}
\newacro{DDP}{Differential Dynamic Programming}
\newacro{COM}{center of mass}
\newacro{COP}{center of pressure}
\newacro{NLP}{nonlinear program}
\newacro{MLP}{Multilayer Perceptron}
\newacro{SLQ}{Sequential Linear-Quadratic}
\newacro{HAA}{hip abduction adduction}
\newacro{AD}{automatic differentiation}
\newacro{HJB}{Hamilton–Jacobi–Bellman}
\newacro{BC}{Behavioral Cloning}
\newacro{IRL}{Inverse Reinforcement Learning}
\newacro{IL}{Imitation Learning}
\newacro{RL}{Reinforcement Learning}
\newacro{MEN}{mixture-of-experts network}
\newacro{MILE}{mixture of implicitly localized experts}
\newacro{MELE}{mixture of explicitly localized experts}
\newacro{DL}{Deep Learning}
\newacro{WBC}{whole-body controller}
\newacro{MDP}{Markov Decision Process}
\newacro{RNN}{Recurrent neural network}
\newacro{PPO}{Proximal Policy Optimization}

\def\TheTitle{Learning Accurate Whole-body Throwing with High-frequency Residual Policy and Pullback Tube Acceleration}

\title{\LARGE \bf \TheTitle} 
\author{Yuntao Ma$^1$, Yang Liu$^2$, Kaixian Qu$^1$, Marco Hutter$^1$%
    \thanks{This work was supported by Intel Labs, by the Max Planck ETH Center for Learning Systems, the Swiss National Science Foundation (SNSF) through project 166232, 188596, the National Centre of Competence in Research Robotics (NCCR Robotics), and the European Union's Horizon 2020 (grant agreement No.852044 and No.101016970). Moreover, this work has been conducted as part of ANYmal Research, a community to advance legged robotics.}%
    \thanks{$^1$ Robotic Systems Lab, ETH Z\"u{}rich, Switzerland.}%
    \thanks{$^2$ Yang Liu is with the Swiss Federal Institute of Technology Lausanne (EPFL), Switzerland. Email: {\tt\footnotesize yangliudh@gmail.com}}%
    \thanks{Corresponding email: {\tt\footnotesize mayuntao94@gmail.com}}%
}

\definecolor{yang}{rgb}{0.54, 0.17, 0.89} 
\begin{document}

\maketitle
%
\thispagestyle{empty}
\pagestyle{empty}
%
%
\begin{abstract}

Throwing is a fundamental skill that enables robots to manipulate objects in ways that extend beyond the reach of their arms. We present a control framework that combines learning and model-based control for prehensile whole-body throwing with legged mobile manipulators. Our framework consists of three components: a nominal tracking policy for the end-effector, a high-frequency residual policy to enhance tracking accuracy, and an optimization-based module to improve end-effector acceleration control. The proposed controller achieved the average of \unit[0.28]{m} landing error when throwing at targets located \unit[6]{m} away. Furthermore, in a comparative study with university students, the system achieved a velocity tracking error of \unit[0.398]{m/s} and a success rate of 56.8\%, hitting small targets randomly placed at distances of \unit[3-5]{m} while throwing at a specified speed of \unit[6]{m/s}. In contrast, humans have a success rate of only 15.2\%. This work provides an early demonstration of prehensile throwing with quantified accuracy on hardware, contributing to progress in dynamic whole-body manipulation. A video summarizing the proposed method and the hardware tests is available at \href{https://youtu.be/3ysgbN6Ca8A}{https://youtu.be/3ysgbN6Ca8A}.
\end{abstract}

%
%

\section{Introduction}\label{sec:introduction}
\begin{figure}[ht]
    \centering
    \includegraphics[width=0.48\textwidth]{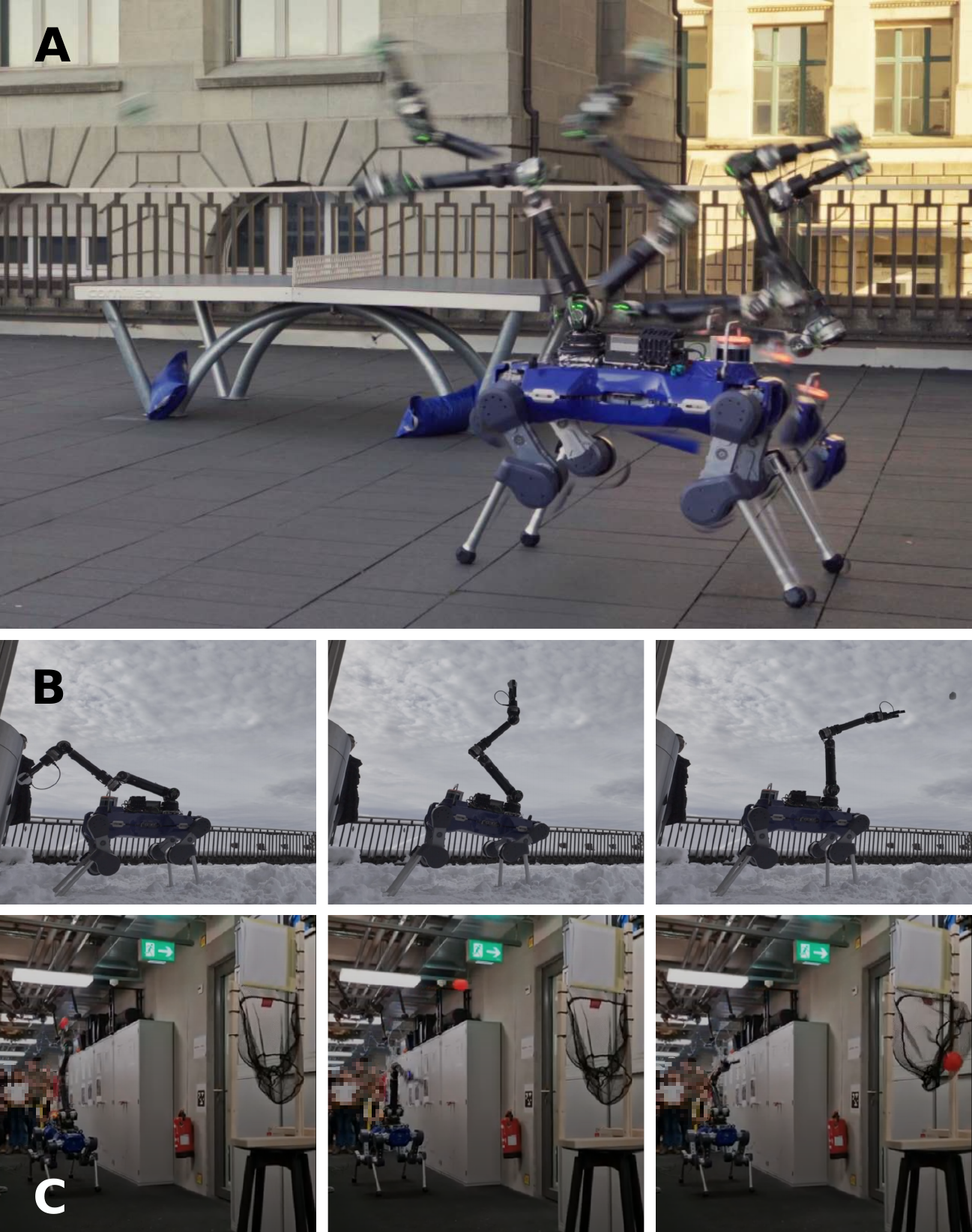}
    \caption{\textbf{The robot performing prehensile whole-body throwing with varying velocities and object properties. (A)} A gift box. \textbf{(B)} A snowball. \textbf{(C)} A floorball. The target positions for the throws are identified using AprilTags, and the EE throwing states are calculated based on a user-specified velocity component along the $x$-axis, aligned with the robot's forward direction.}
    \label{fig:collage}
    
\end{figure}

Legged robots capable of performing whole-body dynamic and high-precision manipulation tasks are essential for advancing applications such as delivery automation, disaster response, and dynamic object handling. By imparting motion to an object, throwing can be used to achieve tasks such as delivering items to distant locations, clearing obstacles, or even enabling dynamic interactions in cluttered environments. Prehensile throwing, where the object is grasped and released mid-motion, extends throwing capabilities by handling irregularly shaped objects and enabling diverse postures. However, it represents a particularly challenging class of throwing that requires precise control of the robot’s end-effector (EE) state and accurate timing of the release.

One of the core challenges in prehensile throwing is the uncertainty in release timing. Variations in gripper-object interactions, such as differences in surface friction or material deformation, can significantly affect the exact moment when the object is released. These uncertainties must be managed to ensure consistent and accurate throws.

Another major difficulty lies in achieving accurate whole-body EE state tracking in highly dynamic scenarios. During a throw, the robot must coordinate its joints and body motion to maintain precise control of the end-effector, even under rapid accelerations and model mismatch. This is particularly challenging for legged mobile manipulators, where the motion of the legs adds additional complexity.

Recent work has explored learning-based approaches to tackle throwing tasks. For instance, Zeng et al. ~\cite{zeng2020tossingbot} and Werner et al.~\cite{werner2024dynamic} have demonstrated throwing capabilities using stationary robotic arms, achieving impressive results for diverse sets of objects. However, these approaches are limited to fixed-base systems. More recent works~\cite{munn2024whole, ha2024umi} introduced whole-body approaches for throwing using legged robots, but these works lacked detailed reporting on accuracy metrics on their hardware. Model-based planning have also been employed in dexterous robotic throwing~\cite{liu2024tube,liu2022solution}. While these methods provide strong theoretical guarantees, they have yet to generalize to whole-body systems.

In this work, we address these challenges by formulating prehensile throwing as a whole-body EE velocity tracking problem. Using reinforcement learning (RL), we train a policy to accurately track timed end-effector trajectories during the throwing interval while accounting for uncertainties in the throwing process. To improve accuracy further, we introduce a high-frequency residual policy that refines the nominal policy’s output and a pullback tube acceleration optimizer module that generates corrective motion commands based on real-time throwing errors. This integrated approach enables robust and accurate prehensile throwing under highly dynamic conditions. 

We summarize our contributions as follows:

$\bullet$ A learning-based control framework for consecutive whole-body prehensile throwing with legged mobile manipulators.

$\bullet$ A high-frequency residual policy for improved EE state tracking during the throw interval.

$\bullet$ An optimization-based module to compute the desired EE acceleration to improve the throwing accuracy.

$\bullet$ Extensive hardware and simulation validation of the proposed method.

%

\section{Related Work}

Whole-body loco-manipulation enables legged robots to seamlessly integrate locomotion and manipulation. Earlier work treated them separately with decoupled controllers~\cite{ma2022combining}. However, more recent approaches aim to unify them into a single whole-body policy for better coordination and efficiency~\cite{fu2022deep}. Liu et al.\cite{liu2024visual} introduced VBC, a vision-based framework that integrates high-level planning with a low-level control policy for robust object grasping. Portela et al.\cite{portela2024whole} developed an RL-based controller for precise end-effector pose tracking, achieving high accuracy on rough terrains. In this work, we focus on whole-body throwing, enabling the robot to leverage its base agility to achieve a higher release velocity.

\paragraph{Residual Policy Learning}

A significant amount of research has focused on enhancing the data efficiency of deep reinforcement learning by integrating model-based methods~\cite{mordatch2016combining, nagabandi2018neural, tan2018sim, johannink2019residual}. Beyond this, other approaches aim to refine arbitrary policies, including but not limited to model-based ones, as in Residual Policy Learning~\cite{silver2018residual}. Recent work has shown that residual learning improves the efficiency of first-order policy gradient algorithms for locomotion and perceptive navigation~\cite{luo2024residual}. Our work follows the approach of refining a learning-based nominal policy with a high-frequency residual policy.

\paragraph{Precise Throwing}

Zeng et al.\cite{zeng2020tossingbot} introduced residual physics, a hybrid approach that enhances physics-based throwing models with data-driven corrections, achieving higher accuracy than purely analytical or learned methods. More recently, dynamic throwing has also been demonstrated on a 12-ton multipurpose excavator using reinforcement learning\cite{werner2024dynamic}. In parallel, Liu et al.~\cite{liu2024tube} proposed tube acceleration, a class of constant-acceleration motion in joint space designed to ensure robustness against release uncertainty. In this work, we seamlessly integrate model-based tube acceleration with model-free RL-based motion generation in closed-loop, to enhance the precision of throwing with a complex legged manipulator system.

%

\begin{figure*}[th!]
    \centering
    \includegraphics[width=0.75\textwidth]{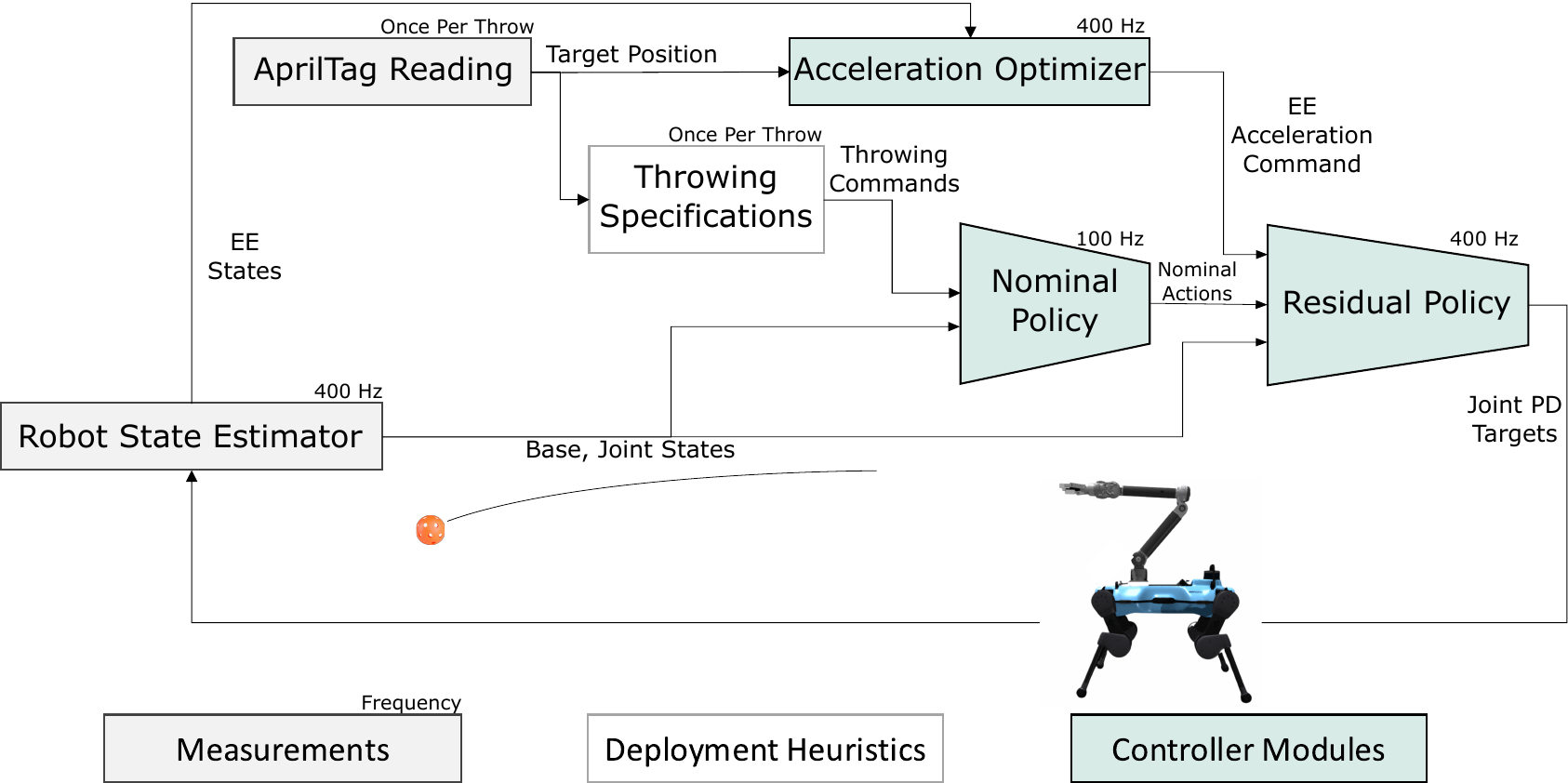}
    \caption{\textbf{The proposed control framework.} consists of a standard whole-body base throwing policy, a residual policy, and an acceleration optimizer. The latter two modules run at the same frequency as the robot state estimation to provide high-frequency state feedback for the throwing motion. During the deployment, the controller acquires the target position from the AprilTag reader and computes the PD target commands for each joint on the robot.}
    \label{fig:pipeline}

\end{figure*}

\section{Method}\label{sec:method}

The proposed throwing controller consists of three components: a baseline whole-body EE state tracking policy (referred to as ``nominal policy''), a residual policy, and a pullback tube acceleration optimizer module, as outlined in Fig.~\ref{fig:pipeline}. This section provides an introduction to the design and implementation of each module.

\subsection{Whole-body EE State Tracking}
Unlike conventional end-to-end formulations, where the policy network \emph{implicitly} generates motion, our method \emph{explicitly} commands a robust throwing motion. The motion remains within the goal manifold~\cite{pekarovskiy2013optimal} of object flight, ensuring robustness to release uncertainty. Therefore, it is essential to ensure precise EE state tracking throughout the release window. To improve the policy’s generalization across different object weights, randomized masses are added to the gripper during training.

The training episodes are structured similarly as outlined in \cite{ma2025badminton}, with multiple throwing targets in each episode. A fixed preparation period of 2 seconds is allocated between the moment the target states are revealed to the policy and the final timestep of the throwing interval. At a control frequency of 100 Hz, the policy observes the robot’s current states, target commands, elapsed time for the current throw, and previous actions. The policy then generates a PD control target for each of the 18 robot joints.

To introduce variability and encourage robust behavior, both position and velocity targets are independently randomized for each throw. At this training stage, a constant velocity reference is specified for each throw to simplify the learning process. The target gripper orientations are those that avoid the gripper fingers blocking the throw velocity direction.

The policy is incentivized to achieve maximum reward by minimizing errors in tracking the specified position, velocity, and orientation targets. Detailed definitions of the observation space and reward functions, including how each component contributes to the overall performance metric, are provided in Table~\ref{tab:rewards}.

\begin{table}[ht]
\caption{\textsc{Training Rewards}} \label{tab:rewards}
\centering
\begin{tabular}{rcll}
\hline
\multicolumn{1}{c}{Reward} & Expression & \multicolumn{1}{c}{Scale} & \multicolumn{1}{c}{Res. Scale} \\ \hline
Termination                & -          & -200                                  & -                                         \\
Torque                     &   $||\bm{\tau}||^2$        & -1e-05                                & -                                         \\
Joint Acc.                 &    $||\ddot{\textbf{q}}||^2$        & -2.5e-07                             & -                                         \\
Collision                  &    -        & -2                                    & -                                         \\
Action Rate                &    $||\textbf{q}^*_k-\textbf{q}^*_{k-1}||^2$        & -0.03                                 & -                                         \\
Torque Limit               &    $|\text{clip(}\bm{\tau}-\bm{\tau}_{\text{lim}}\text{, min=0)}|$        & -0.001                                & -0.001                                    \\
Standing                   &    $||\textbf{q}-\textbf{q}_\text{stand}||^2$, no cmd        & 24                                    & -                                         \\
Current Limit              &    $\text{clip(}I-I_{\text{lim}}\text{, min=0)}$        & -0.1                                  & -0.1                                      \\
EE Pos.           &    ${1}/{(1+||\bm{p}_{EE}^*-\bm{p}_{EE}||^2)}$        & 4000                                  & 4000                                      \\
EE Vel.           &   ${1}/{(1+|v_{EE}^*-v_{EE}|)}$         & 2000                                  & 2000                                      \\
EE Alignment    &     ${1}/{(1+|_{EE}v_x|^2)}$       & 1000                                  & 1000                                      \\
Action Scale               &    $||{\textbf{q}_\text{residual}}||^2$        & -                                     & -1                                        \\ \hline
\end{tabular}
\end{table}

\subsection{High-frequency Residual Policy}

Once the nominal policy is trained, we freeze its weights and train a residual policy to enhance the accuracy of EE tracking and to enable effective EE acceleration tracking. This policy operates at \unit[400]{Hz}, matching the frequency of the robot’s state estimation—the highest rate that provides real-time feedback actions. In addition to the nominal policy observations, the residual policy observes the current control decimation and the EE acceleration target in the vertical direction. The residual policy outputs residual arm joint position offsets that are then added to the nominal policy action outputs for the corresponding joints.

The residual policy is trained using a combination of rewards: the nominal policy’s task rewards with modified position and velocity references based on the acceleration, and an additional reward for minimizing the scale of its actions. The action scale reward encourages the residual policy to improve tracking performance without introducing significant deviations from the nominal policy’s behavior.

Since the residual policy operates on a faster time scale than the nominal policy and its training prioritizes exploitation over exploration, it requires a separate set of training hyperparameters optimized for its higher-frequency dynamics. These parameters are detailed in Table~\ref{tab:hyperparams}.

\begin{table}[ht]
\caption{\textsc{Training Hyperparameters}} \label{tab:hyperparams}
\centering
\begin{tabular}{rcc}
\hline
\multicolumn{1}{c}{Hyperparameter} & Nominal Policy & Residual Policy \\ \hline
Control Freq.                      & 100 Hz      & 400 Hz          \\
Action Scale                       & 0.5         & 0.2             \\
Sim. Steps Per Iter.                     & 96          & 384             \\
Entropy Coeff.                     & 0.0016      & 0.035           \\ \hline
\end{tabular}
\end{table}

\subsection{Robust Throwing with Pullback Tube Acceleration}
Besides tracking error, another challenge of accurate high-speed prehensile throwing arises from the amplified effects on release uncertainty, which already influences the success rate for low-speed throwing ($\sim$1.5 m throws) significantly~\cite{monastirsky2022learning, liu2024tube}. While one can applying domain randomization on release timing to train an end-to-end robust RL policy, Monastirsky et al.~\cite{monastirsky2022learning} reported that direct deployment of such trained policy resulted in highly dangerous movement for fixed-base manipulator. This raises our concerns to apply such strategy on a 58-kg legged-manipulator for high-speed throwing. Therefore, in this work, we adopt the strategy proposed by Liu and Billard~\cite{liu2024tube} that explicitly \emph{synthesize} feasible and safe robust throwing motion against release uncertainty. In~\cite{liu2024tube}, the EE's state during the release window (with duration 100 ms) is driven by a constant Tube Acceleration, designed to be \emph{traversing} in the backward reachable tube (BRT) of valid throwing states, such that the landing position is agnostic to release timing.

Despite being highly effective for robust dexterous throwing with fixed manipulators, ~\cite{liu2024tube} assumes an accurate tracking controller is readily available, and hence the robust release motion is generated given the \emph{planned} nominal throwing state before executing the throwing motion. However, such \emph{open-loop} architecture is insufficient in our setup. In particular, despite the improved tracking accuracy thanks to the high-speed residual policy, the remaining tracking error could make a synthesized release motion associated with the \emph{nominal} EE throwing state ineffective for the \emph{actual} EE state. Furthermore,~\cite{liu2024tube} assumes that the EE state already resides in the BRT upon entering the release phase, while in our case, the tracking error might lead to an invalid EE state when starting the release phase.

To address the limitations of the original Tube Acceleration, in this work, we made an insightful observation: a \emph{closed-loop} Tube Acceleration controller based on \emph{current} EE state will \emph{pullback} the EE state into the BRT, effectively making the BRT an attracting invariant set of the closed-loop dynamical system.

\subsubsection{Backward Reachable Tube}\label{subsec:BRT}
In the throwing plane, the object flying state is denoted as $\xi = [r, z, \dot{r}, \dot{z}]^\top \in \mathbb{R}^4$. The flying dynamics is described by a first-order differential equation $\dot{\xi} = f_{fly}(\xi)$. The flying trajectory of $f_{fly}$ starting from detach state $\xi_{d}$ are denoted as $\zeta_{f_{fly}, \xi_{d}}(t):[0, +\infty] \rightarrow \mathbb{R}^4$. We assume that a user has provided the robot with a landing target set $\mathcal{X} \subset \mathbb{R}^4$, which describes the allowed landing position slack and the range of allowed landing velocities.

For a flying trajectory that enters the landing target set, any state on this trajectory segment is a valid throwing configuration. Therefore, by aggregating all the trajectories that eventually enter the landing target set, we obtain the set of valid throwing configurations, which we call the backward reachable tube (BRT). Mathematically, the BRT is defined as:
\begin{align*}
    \mathcal{G}(f_{fly}, \mathcal{X}) = \{\xi_{d} \, | \, \exists \, t \geq 0, \, \zeta_{f_{fly}, \xi_d}(t) \in \mathcal{X}\}.
\end{align*}
Given a connected target set $\mathcal{X} \subset \mathbb{R}^4$, the BRT $\mathcal{G}$ associated with a smooth continuous flying dynamics $f_{fly}$ is also a connected set in $\mathbb{R}^4$ without any isolated regions (or `holes') (Th. 3.5 \cite{khalil2002nonlinear}). 
As a result, BRT $\mathcal{G}$ is defined in a topological space, with well-defined topological concepts such as boundaries and interiors. 

For projectile dynamics, the BRT can be analytically described by the object's flying flowmap $\Phi$ that maps from release state $\xi_d$ and target height $z_{land}$ to the horizontal landing position $r_{land}$, i.e.,
\begin{align*}
   r_{land} &=  \Phi(\bm{p}_d, \bm{v}_d, z_{land}) = r_d+\dot{r}_d\Delta t_{fly} \\
&=r_d+\dot{r}_d\frac{\dot{z}_d+\sqrt{\dot{z}_d^2+2G(z_d-z_{land})}}{G}
\end{align*}
where $(\bm{p}_d, \bm{v}_d)$ is the EE state upon detachment, $G = 9.81 \text{ m}/\text{s}^2$ is the gravitational acceleration. 
It is worth noting that although the analytical expression of the flying flowmap is available for projectile motion—where the object is solely influenced by gravitational force, such analytical solutions are generally not available for nonlinear flying dynamics. This holds true even for object flying dynamics with a simple quadratic air-drag model, as discussed in the review by Lubarda et al.~\cite{lubarda2022review}. In this case, neural implicit representation of the BRT via Neural Event ODE~\cite{chen2021learning} can be employed, as proposed in~\cite{liu2024tube}.

\subsubsection{Pullback Tube Acceleration}
Due to unrepeatable gripper opening motion (arising from the screw drive fingers) and the object deformation upon grasping, the \emph{detach time} after giving gripper opening command cannot be determined \emph{a priori}, but only known to be between $50$-$100$ ms ($50$ ms gripper dwell time plus unknown normal force vanishing time). Therefore, we try to drive the EE motion to enter and stay inside the BRT of the target during the $100$ ms release window via \emph{commanding} appropriate acceleration, we call Pullback Tube Acceleration. Note that if the EE's state is outside the BRT when giving the gripper opening command, the $50$ms gripper dwell time provides the control authority to bring the EE's state back to the BRT.  

The \emph{realtime} Pullback Tube Acceleration is generated by the following convex optimization problem,  \textcolor{blue}{blue} variables are \emph{decision variables} induced by tube acceleration \textcolor{blue}{$\bm{a}_{tube}$}, while black variables can be viewed as \emph{parameters} of the program and hence are treated as fixed in the solver:
\begin{align*}
\textbf{Problem} &\textbf{\textit{ Tube}} \textbf{\textit{{-CVX}}}\\
\text { min  } & \left|\textcolor{blue}{r_{land}} - r_{target}\right |^2 \quad \\
\text { s.t.: } 
& \textcolor{black}{\bm{p}_{T} = \bm{p}_{EE} + T \bm{v}_{EE}}\\
& \textcolor{blue}{\bm{v}_{T}} = \bm{v}_{EE} + T \textcolor{blue}{\bm{a}_{tube}}\\
& \textcolor{blue}{\dot{r}_T} = \|\textcolor{blue}{\bm{v}_{T,xy}}\|_2\\
& \textcolor{blue}{\dot{z}_T} = \textcolor{blue}{\bm{v}_{T,z}} \\
&\dot{r}=\|\bm{v}_{EE, xy}\|_2\\
&\dot{z} = \bm{v}_{EE, z}\\
& r_{land}^0 = \Phi(\bm{p}_{T}, \bm{v}_{EE}, z_{land}) \\
& \textcolor{blue}{r_{land}} = r_{land}^0+ \left[\frac{\partial r_{land}^0}{\partial \dot{r}}, \frac{\partial r_{land}^0}{\partial \dot{z}}\right]^\top\left[\begin{array}{c}
\textcolor{blue}{\dot{r}_{T}}-\dot{r} \\
\textcolor{blue}{\dot{z}_{T}}-\dot{z}
\end{array}\right]\\
& \bm{v}_{\min } \leq \textcolor{blue}{\bm{v}_{T}}  \leq \bm{v}_{\max }\\
& \bm{a}_{\min } \leq \textcolor{blue}{\bm{a}_{tube}}  \leq \bm{a}_{\max }
\label{opt-convex}
\end{align*}
\noindent where $(\bm{p}_{EE},\bm{v}_{EE})$ is the \emph{online measurement} of the EE state, $T$ is the time left until the terminal of the $100$ms release window. In the above formulation, all the equality constraints are linear, and all the inequality constraints are polytopic, resulting in a convex program~\cite{boyd2004convex}. Liu and Billard~\cite{liu2024tube} proved and experimentally demonstrated that this convex program is a tight relaxation of the primal nonconvex robust throwing problem. In this work, the above formulation detail readily follows Disciplined Parametrized Programming (DPP)~\cite{agrawal2019differentiable} and hence natively supported by CVXPYgen~\cite{schaller2022embedded}, resulting an average solving time of $0.4$ms, 75$\times$ speed up compared to the $30$ms solving time of the original DPP-incompatible formulation in~\cite{liu2024tube}. Consequently, the Pullback Tube Acceleration is capable of running in closed-loop with over 1kHz frequency to robustify throwing motion.

\subsection{Training and Implementation}

The policy training is conducted in \text{legged\_gym}\cite{rudin2022learning} on a workstation with a NVIDIA RTX3080Ti with Proximal Policy Optimization~\cite{schulman2017proximal}. The nominal policy and residual policy training require 4,500 and 1,200 iterations, corresponding to \unit[2,457.6]{h} and \unit[655.4]{h} of simulation time, respectively.  Standard sim-to-real techniques including actuator networks~\cite{hwangbo2019learning}, domain randomization~\cite{tobin2017domain}, observation noises, and symmetry augmentation~\cite{mittal2024symmetry} are applied to both training stages to improve policy robustness and visual appeal.

The policy is deployed on an ANYmal robot \cite{hutter2016anymal} equipped with a Duatic DynaArm and a Robotiq 2f140 gripper. Onboard state estimation running at \unit[400]{Hz} provides the observations for the policy. The nominal policy is decimated to operate at \unit[100]{Hz}, while the residual policy and the pullback tube acceleration run at its full rate of \unit[400]{Hz}.

For hardware experiments, throwing targets are identified using AprilTag \cite{olson2011apriltag}. The nominal throwing velocity is determined by computing a parabolic flight trajectory for the object based on the release end-effector position, neglecting air drag to simplify calculations. The pullback tube is implemented as an asynchronous rosnode to provide the latest EE acceleration target.

\section{Results}

We deployed our proposed control framework to perform whole-body prehensile throwing tasks with a variety of object types across diverse scenarios. The system demonstrated repeated success in throwing objects such as gift boxes, snowballs, and floorballs, both indoors and outdoors, as shown in Fig.~\ref{fig:collage} and the supplementary video. Notably, throwing snowballs was particularly challenging due to their highly variable properties, including slippage, inconsistent mass distribution, and deformable shapes. Despite these challenges, the system successfully threw snowballs to targets located \unit[5-7]{m} away with consistent accuracy.

Qualitatively, the controller effectively coordinated all limbs to achieve throws of varying distances, heights, lateral directions, and specified speeds, leveraging the combined contributions of the baseline tracking policy, residual policy, and optimization module.

Note that some experimental results were conducted using partial implementations of the proposed method or earlier training checkpoints due to timeline constraints. For example, the snowball throwing tests and the comparison of accuracy with human participants were performed under such conditions.

\subsection{Comparison with Human Throwing}

In a public demo (Fig.~\ref{fig:collage}C), we compared the robot’s throwing accuracy with that of 25 student participants. The task involved throwing a standard floorball at a target measuring 25 × 28 cm, randomly placed 3-4 meters away in a corridor. The robot was configured to throw with a fixed horizontal velocity of \unit[6]{m/s}, while participants were allowed to throw in any manner they preferred. The participant and the robot took turns to have five throws at the target each round. Human participants collectively scored 19/125, with the highest individual scores being 3/5. In contrast, the robot achieved a score of 71/125, significantly outperforming the human participants.

\subsection{Throwing Accuracy}

To evaluate the overall throwing accuracy of the proposed controller, we conducted experiments targeting ground locations at distances of 4 and 6 meters in the robot’s forward direction, each with lateral offsets of -0.5 meters and 0.5 meters. The nominal throwing velocities for each target were computed for the forward and vertical directions based on the specified release position.

We measured the landing position error magnitude for each throw and compared the performance of the proposed controller with the nominal policy, which does not include the residual policy or the pullback tube acceleration optimizer. The experiment consisted of 10 throws per target location, resulting in a total of 40 throws. The same experiment is repeated with the nominal policy for comparison to evaluate the improvement from the residual policy and pullback tube acceleration.

The results demonstrate that the mean landing error for the 4-meter target using the proposed controller is \unit[0.429]{m}, while the error for the 6-meter target is \unit[0.276]{m}. In comparison, the nominal policy produced errors of \unit[0.710]{m} and \unit[0.685]{m} for the 4-meter and 6-meter targets, respectively. These results highlight an average of 49.5\% accuracy improvement achieved by integrating the residual policy and tube acceleration optimizer into the control framework.

\subsection{Base Motion Contribution}

To evaluate the contribution of the legged base to the throwing motion, we analyzed the torque, angular impulse, and the corresponding power and work exerted by the base on the arm prior to the throw. Torque was computed using inverse dynamics with Pinocchio~\cite{carpentier2019pinocchio} based on the arm base pose, as well as the joint position, velocity, and acceleration trajectories. Joint positions and velocities were obtained from onboard state estimation, while joint accelerations were calculated via finite differences of the velocity data and subsequently low-pass filtered to reduce noise. Angular impulse and torsional power were then derived from the computed torque and the arm base velocity. The data were recorded from five throws with the horizontal velocity of \unit[10]{m/s}.

\begin{figure}[ht]
    \centering
    \includegraphics[width=0.48\textwidth]{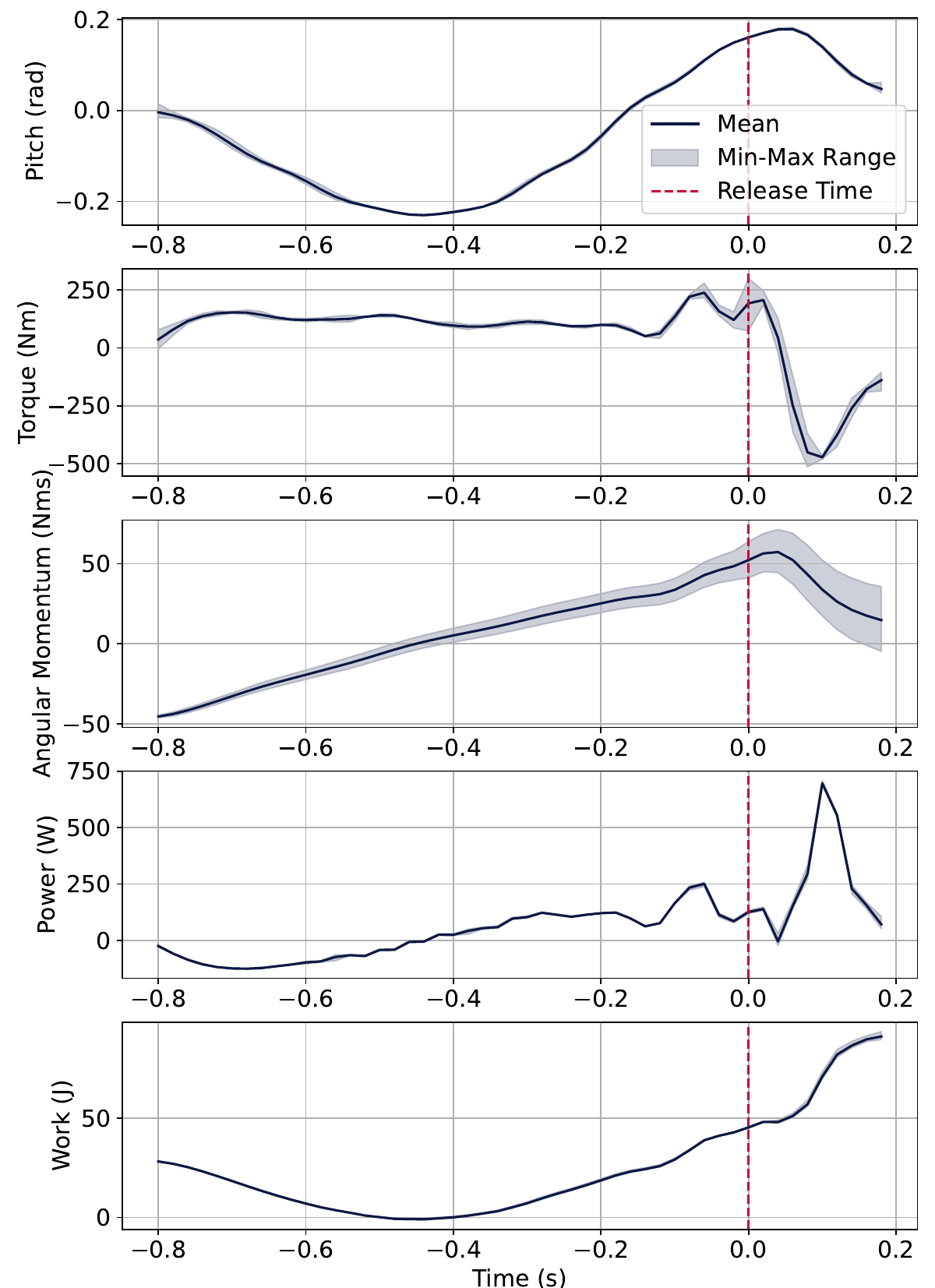}
    \caption{\textbf{Base motion contributions to throws.} This figure illustrates the role of base motion in the throwing process by analyzing the base pitch angle, pitching torque, pitching angular momentum, power, and cumulative work. The angular momentum and cumulative work are calculated starting from the moment the base pitch rate turns positive. Data is derived from robot state estimation during hardware experiments for throws with a horizontal velocity of \unit[10]{m/s}.}
    \label{fig:base_motion}
    
\end{figure}

Fig.~\ref{fig:base_motion} presents these values, starting from the moment the gripper begins its forward motion, indicating that the base contributed \unit[51.7]{Nms} of angular momentum and \unit[46.3]{J} of work prior to the gripper’s release. To further contextualize these contributions, we compared the angular impulse with that of a tabletop manipulator executing the same joint command trajectory. The analysis indicated that the legged base provided an average of 53.4\% higher angular impulse than the table-top manipulator, showing that the base enhances the overall throwing dynamics. Notably, we observe that the base stays largely stationary at low throwing velocities, with base tilt emerging as velocity increases—showcasing the advantages of using a legged mobile manipulator for dynamic throwing (See supplementary video). 

\subsection{Acceleration Pullback under Ideal Tracking}
In this subsection, we assess the effect of real-time Pullback Tube Acceleration in robustifying stochastic throwing systems. The batch stochastic experiment setup is as follows: $3$ nominal EE heights, $5\times4=20$ nominal EE velocities are perturbed by $5\times5=25$ error ratios, resulting in a mesh of $3\times20\times25=1500$ initial EE states upon entering the release window (after giving the gripper opening command). The nominal landing positions are computed via projectile dynamics based on the nominal EE states $(z, \dot{r}, \dot{z})$ with a zero landing height. Table~\ref{tab:tube_conditions} summarizes the mesh of experiment conditions.

\begin{table}[ht!]
    \centering
    \caption{\small \textsc{Pullback Tube Acceleration Experimental Conditions}}
\label{tab:tube_conditions}
    \begin{tabular}{lc}
        \toprule
        \textbf{Parameter} & \textbf{Values} \\ 
        \midrule
        Nominal EE height $z$ [m] & $\{0.5, 1.0, 1.5\}$ \\
        Nominal EE $\dot{r}$ [m/s] & $\{5.0, 6.0, 7.0, 8.0, 9.0\}$ \\
        Nominal EE $\dot{z}$ [m/s] & $\{1.0, 2.0, 3.0, 4.0\}$ \\
        $\dot{r}$ error ratios & $\{-10\%, -5\%, 0\%, 5\%, 10\%\}$ \\
        $\dot{z}$ error ratios & $\{-10\%, -5\%, 0\%, 5\%, 10\%\}$ \\
        \bottomrule
    \end{tabular}
\end{table}

\begin{figure}[t!]
    \centering
    \includegraphics[width=0.5\textwidth]{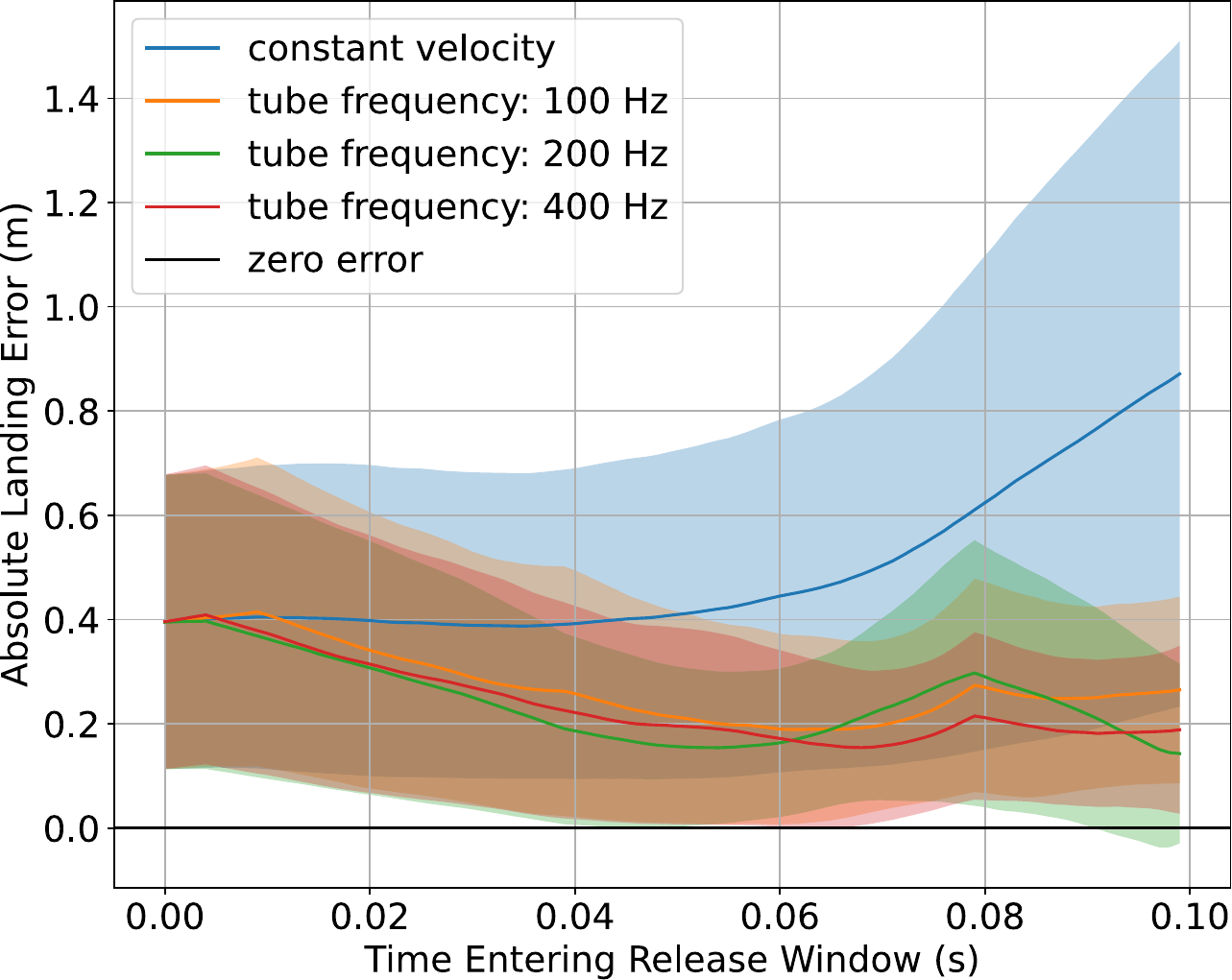}
    \caption{\small Robustifying effect of Pullback Tube Acceleration on stochastic systems.}
\label{fig:profiling_tube_acc}
\end{figure}

At each timestep, the Pullback Tube Acceleration is generated with the simulated EE state measurement (assuming no measurement error). After computing the current Pullback Tube Acceleration command, 
Gaussian noise with a standard deviation of 2 $\text{m}/\text{s}^2$  is applied to the acceleration to simulate actuation error. The simulation runs with a timestep of 1 ms and is repeated for 5 random seeds. The results are summarized in Fig.\ref{fig:profiling_tube_acc} and Table~\ref{tab:profiling_tube_acc}.

\begin{table}[ht!]
    \centering
    \caption{\small \textsc{Pullback Tube Acceleration profiling}}
\label{tab:profiling_tube_acc}
    \begin{tabular}{lc}
        \toprule
        {Release Motion Command} & Max. Landing Error (cm) \\ 
        \midrule
        Constant Velocity & 96.8(±57.9)\\
        Pullback Tube Acceleration (100Hz) & 38.1(±57.9)\\
        Pullback Tube Acceleration (200Hz) & 34.9(±23.6)\\
        Pullback Tube Acceleration (400Hz) & \textbf{31.1(±17.7)}\\
        \bottomrule
    \end{tabular}
\end{table}

Fig.~\ref{fig:profiling_tube_acc} illustrates that due to perturbations, upon entering the release window, there is an mean absolute error (MAE) of 40 cm in landing position. For the `Constant Velocity Command' without a robustifying release motion, the error continues to increase, reaching an average of 90 cm. On the other hand, for systems applying real-time Pullback Tube Acceleration, the initial error is significantly reduced within the first 50 ms of the gripper dwell time and remains small ($\sim20$ cm) throughout the 50–100 ms window, which represents the potential detach instant.

Table~\ref{tab:profiling_tube_acc} reports the quantitative effect of Pullback Tube Acceleration. We evaluate the MAE and standard deviation of the maximum landing error in the 50–100 ms possible detach window, averaged over nominal EE states, initial EE velocity perturbations, and acceleration noise random seeds. We observe a monotonic decrease in error as control frequency increases, demonstrating the necessity of real-time robust motion synthesis.

\subsection{Ablation Studies}

\begin{table*}[ht!]
\caption{\textsc{Ablation Studies}} \label{tab:ablation}
\begin{tabular}{rcccccc}
\hline
\multicolumn{1}{l}{}                    & Pos. Tracking & Vel. Tracking &  Orient. Alignment & \begin{tabular}[c]{@{}c@{}}Landing Err.\\ L1 (m)\end{tabular} & \begin{tabular}[c]{@{}c@{}}Landing Err.\\ MSE. (m$^2$)\end{tabular} & Success Rate \\ \hline
Nominal Policy                      &      0.0693         &   0.2506            &            0.0054       &        0.2530                                                         &      0.1658                                                            &                63.3\%                 \\
Residual 100 Hz                         &        0.0466       &    0.2269           &    0.0053               &         0.2139                                                        &          0.0692                                                        &     13.9\%         \\
Residual 400 Hz Original Hyp.              &     0.0478          &    0.2105           &         0.0048          &     0.2213                                                            &   0.0701                                                               &        65.5\%      \\
Residual 400 Hz              &        0.0490       &    0.2070           &            0.0050       &        0.2156                                                         &       0.0728                                                           &   69.3\%          \\
Residual 400 Hz + Tube Acc. (ours) & \textbf{-}             & \textbf{-}             & \textbf{-}                 &      \textbf{0.2023}                                                           &           \textbf{0.0587}                                                       &     \textbf{70.8\%}         \\ \hline
\end{tabular}
\vspace{-2em}
\end{table*}

To evaluate the contributions of individual components in the proposed control framework, we conducted an ablation study comparing the performance of different controller variations. The tested variations include:

\begin{enumerate}
    \item The nominal policy.
    \item A residual policy running at 100 Hz.
    \item A residual policy running at 400 Hz with the same training hyperparameters as the nominal policy.
    \item A tuned residual policy running at 400 Hz.
    \item The full proposed method, combining the tuned 400 Hz residual policy with the pullback tube acceleration optimizer.
\end{enumerate}

The evaluation was performed in simulation, assessing both EE tracking error and the ability to throw an object accurately toward a ground target. The target was positioned 7 meters directly in front of the robot, and the robot was commanded to throw with a horizontal velocity of \unit[7]{m/s}. The nominal vertical velocity was computed to be \unit[3.205]{m/s} based on the specified release position. Variations without the pullback tube acceleration optimizer were instructed to track this constant velocity, while the proposed method tracked varying EE velocities during the throw based on the computed acceleration. The throw is counted as successful if the landing position is within \unit[0.25]{m} from the ground target.

Table~\ref{tab:ablation} presents the results of the evaluation, conducted over 500 simulated trials. The results indicate that the tuned 400 Hz residual policy consistently outperformed the other variations in EE tracking performance. Moreover, the landing accuracy of the proposed method, which includes the pullback tube acceleration optimizer, was the best among all variations. The proposed method achieved a landing position improvement of 6.17\% over the variation without the pullback tube acceleration optimizer and 20.04\% over the nominal policy.

\subsection{EE Velocity Tracking}

\begin{figure}
    \centering
    \begin{subfigure}{0.492\columnwidth}
        \centering
        \includegraphics[width=\textwidth]{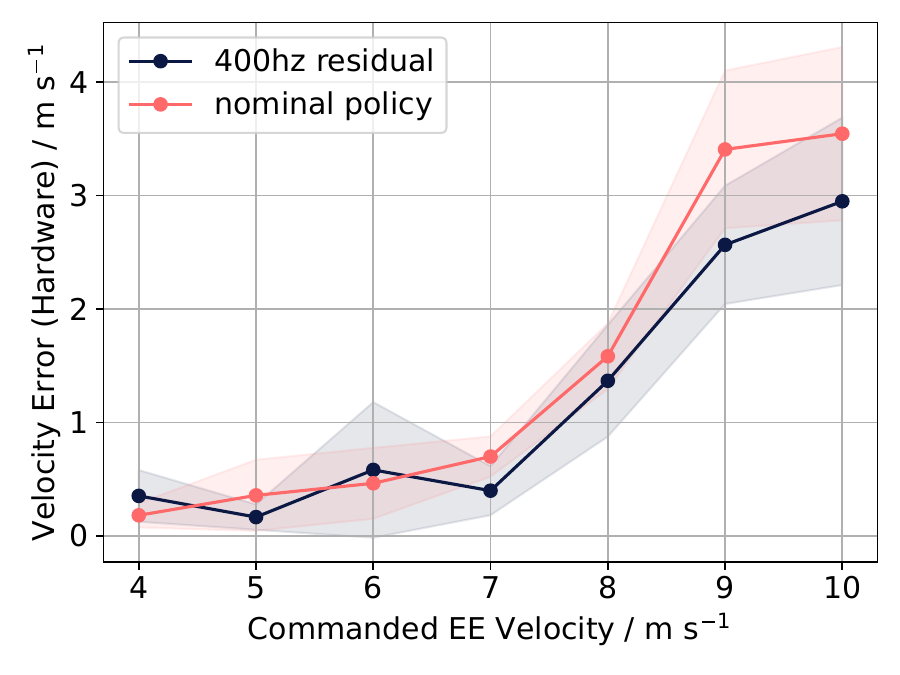}
        \caption{}
        \label{fig:swing_hw}
    \end{subfigure}
    \begin{subfigure}{0.492\columnwidth}
        \centering
        \includegraphics[width=\textwidth]{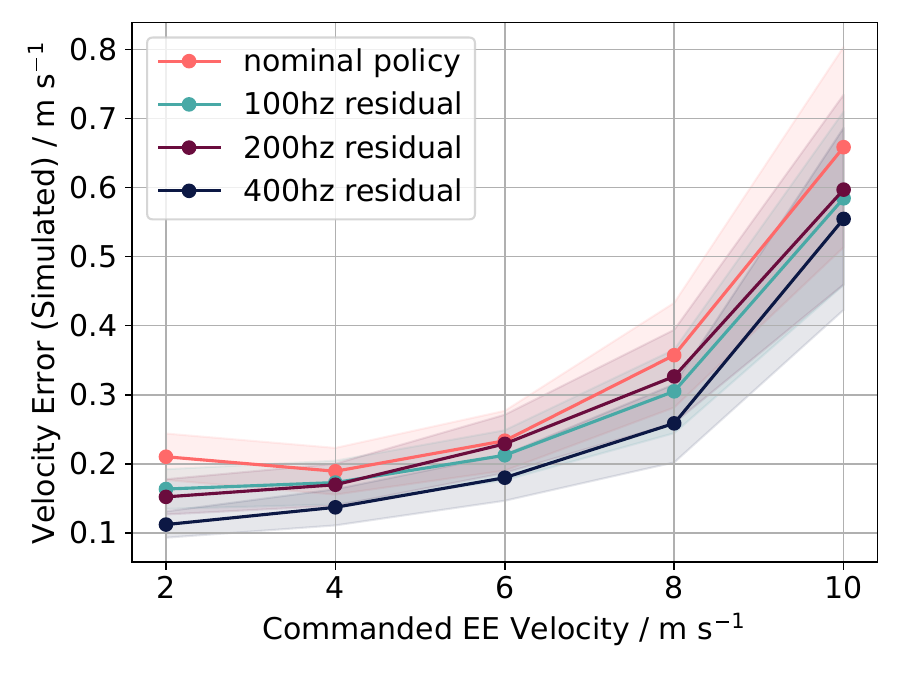}
        \caption{}
        \label{fig:swing_sim}
    \end{subfigure}
    \caption{\textbf{Comparison of residual policy effects in hardware and simulation.} 
    (a) EE velocity tracking improvement on hardware with the residual policy. The residual policy enhances tracking, especially at higher commanded end-effector velocities. 
    (b) Effectiveness of residual policies with various frequencies in simulation. All tested frequencies improve tracking over the baseline, with the 400 Hz residual policy achieving the best performance.}
    \label{fig:swing_combined}
\end{figure}

To evaluate the impact of the high-frequency residual policy during the hardware deployment, we analyzed EE velocity tracking accuracy during throws at various commanded velocities. Fig.~\ref{fig:swing_hw} shows the comparison between the nominal policy alone and the combined policy with the residual. The plot presents the mean velocity tracking error during the commanded throw duration of 80 ms. At lower velocities (4-6 m/s), there is no clear difference between the two policies. However, at higher velocities (7-10 m/s), the residual policy provides noticeable improvements, reducing tracking errors. Specifically, during throws at the highest tested speed of \unit[10]{m/s}, the mean velocity error is 16.8\% lower when using the residual policy compared to the nominal policy alone, justifying its inclusion for enhanced performance in high-speed scenarios.

\subsection{Residual Policy at Various Frequencies}

We also compared the performance of the residual policy at different operating frequencies through simulated experiments. The training hyperparameters for each setting are independently tuned for one working day. Fig.~\ref{fig:swing_sim} shows the velocity tracking error over the same interval used in the previous subsection (EE velocity tracking accuracy), with results collected from the best of three seeds for each residual policy setting through 500 simulated throws. All residual policies were trained using the same nominal policy for consistency. The comparison indicated that the 400 Hz residual policy consistently outperformed policies operating at lower frequencies. Notably, even the residual policy running at 100 Hz achieved improvements over the nominal policy, demonstrating the benefits of high-frequency refinement.

However, we acknowledge that all simulated setups performed significantly better than the corresponding hardware results, indicating that the residual policy also suffers from the sim-to-real gap. While whole-body control (WBC) has shown substantial improvements in the transferability of model predictive control (MPC) and trajectory optimization (TO) controllers, our residual policy only provides moderate enhancements. It appears to overfit the dynamics of the training environment rather than learning generalized local feedback. Addressing this limitation could involve improving the training simulation's system dynamics model or enhancing the residual policy's error feedback behavior.

\section{Conclusion} 

We proposed a learning-based control framework for whole-body prehensile throwing with legged manipulators. Our approach integrates a high-frequency residual policy for EE state tracking and an optimization-based module for desired accelerations, improving accuracy. Extensive experiments validated its effectiveness in dynamic manipulation.


Our experiments confirm the framework’s ability to achieve fast and accurate throws, resulting in the average of \unit[0.276]{m} landing error when throwing at targets located \unit[6]{m} away, improving accuracy by 49.5\% over the baseline and significantly outperforming humans in our comparative study. To our knowledge, this is the first reported instance of whole-body prehensile throwing with quantified accuracy on hardware.

Several limitations remain to be addressed. The current residual policy implementation simplifies training by only tracking accelerations in the vertical direction, which restricts the system's overall accuracy. Moreover, while the residual policy improves tracking accuracy, the EE velocity tracking results indicate that its impact is less pronounced compared to the enhancements achieved by reference trackers in TO and MPC. This limitation is partially resolved by the pullback tube acceleration optimizer, as shown in the landing distance comparison. Future work could further explore advanced variations of the residual policy to achieve greater tracking precision. Additionally, the framework assumes that the object's mass is negligible relative to the reflected inertia of the end-effector. Future studies could address this limitation by estimating the inertial properties of objects during the preparation phase of the throws.


\bibliographystyle{IEEEtran}
\bibliography{sources} 
\end{document}